\documentclass[conference]{IEEEtran}
\IEEEoverridecommandlockouts

\usepackage{cite}
\usepackage{amsmath,amssymb,amsfonts}
\usepackage{algorithmic}
\usepackage{graphicx}
\usepackage{algorithm,algorithmic}
\usepackage{hyperref}
\hypersetup{hidelinks=true}
\usepackage{textcomp}
\usepackage{booktabs}
\usepackage{caption}
\usepackage{xcolor}
\usepackage{colortbl}
\def\BibTeX{{\rm B\kern-.05em{\sc i\kern-.025em b}\kern-.08em
    T\kern-.1667em\lower.7ex\hbox{E}\kern-.125emX}}

\newcommand{\red}[1]{{#1}}

\begin{document}

\title{Adapting LLMs for the Medical Domain in Portuguese: A Study on Fine-Tuning and Model Evaluation\\
\thanks{This study was funded by the S\~ao Paulo Research Foundation (FAPESP) grants $2013/07375-0$, $2019/07665-4$, $2023/14427-8$, $2024/00789-8$, and $2024/01336-7$, and the National Council for Scientific and Technological Development (CNPq) grants $308529/2021-9$ and     $400756/2024-2$.\\
This work has been submitted to the IEEE for possible publication. Copyright may be transferred without notice, after which this version may no longer be accessible.}
}

\author{\IEEEauthorblockN{Pedro Henrique Paiola\IEEEauthorrefmark{1},
Gabriel Lino Garcia\IEEEauthorrefmark{1},
João Renato Ribeiro Manesco\IEEEauthorrefmark{1}, \\
Mateus Roder\IEEEauthorrefmark{1}, Douglas Rodrigues\IEEEauthorrefmark{1} and
João Paulo Papa\IEEEauthorrefmark{1}}
\IEEEauthorblockA{\textit{School of Sciences} \\
\textit{São Paulo State University (UNESP)}\\
Bauru - SP, Brazil \\
Email: \{pedro.paiola, gabriel.lino, joao.r.manesco, mateus.roder, d.rodrigues, joao.papa\}@unesp.br}
}


\maketitle

\begin{abstract}
This study evaluates the performance of large language models (LLMs) as medical agents in Portuguese, aiming to develop a reliable and relevant virtual assistant for healthcare professionals. The HealthCareMagic-100k-en and MedQuAD datasets, translated from English using GPT-3.5, were used to fine-tune the ChatBode-7B model using the PEFT-QLoRA method. The InternLM2 model, with initial training on medical data, presented the best overall performance, with high precision and adequacy in metrics such as accuracy, completeness and safety. However, DrBode models, derived from ChatBode, exhibited a phenomenon of catastrophic forgetting of acquired medical knowledge. Despite this, these models performed frequently or even better in aspects such as grammaticality and coherence. A significant challenge was low inter-rater agreement, highlighting the need for more robust assessment protocols. This work paves the way for future research, such as evaluating multilingual models specific to the medical field, improving the quality of training data, and developing more consistent evaluation methodologies for the medical field.
\end{abstract}


\section{Introduction}
\label{sec:introduction}
\IEEEPARstart{L}{arge} Language Models (LLMs) have revolutionized various domains by showcasing their ability to comprehend and generate human-like text. Their applications encompass various fields such as natural language processing, translation, and conversational agents~\cite{Chowdhery:2024,Du:2022,Brown:2020}. However, their potential in the medical domain, where precision and reliability are paramount, has only recently begun to be fully explored.

The integration of LLMs into the medical field represents a significant technological advancement. These models, trained on extensive datasets, employ deep neural network architectures to process and produce natural language text with human-like comprehension. They have demonstrated remarkable capabilities in understanding medical terminology, synthesizing complex information, and assisting in various clinical and administrative tasks.

Medical data's intricate and extensive nature presents significant challenges for healthcare professionals, from analyzing electronic health records to interpreting scientific articles. In this demanding context, LLMs emerge as potent tools that can effectively organize, interpret, and apply information, enhancing the overall workflow and accuracy in medical settings. For instance, in diagnostic assistance, they can analyze symptoms and recommend potential diagnoses, and in research, they assist in literature reviews and trend identification. Additionally, these models can automate hospital administration tasks such as report generation and clinical documentation, thus freeing up valuable time for healthcare providers.

Despite their potential, LLMs encounter challenges related to accuracy, data privacy, algorithmic bias, and regulatory compliance, necessitating careful integration into clinical workflows to maximize their benefits while mitigating risks~\cite{Lee:2023}.

BioBERT~\cite{Lee:2020} represents one of the initial Language Models designed specifically for the biomedical sphere. Derived from BERT, BioBERT underwent training on extensive biomedical literature from sources like PubMed and PMC. Its primary objective is to enhance NLP tasks within the medical domain. It is showcased that BioBERT outperforms other pre-trained models in tasks such as named entity recognition, entity relationships, and question answering, owing to its enhanced comprehension of biomedical terminology and context.

PubMedBERT~\cite{Gu:2021} is a specialized model trained exclusively on biomedical texts from PubMed. This model addresses the unique challenges of understanding and generating natural language within the biomedical context. PubMedBERT demonstrates improved accuracy in various biomedical NLP tasks, including text classification and information extraction, compared to models trained on more general corpora.

Singhal et al.~\cite{Singhal:2023} employed the Flan-PaLM 540B to encode clinical knowledge on various benchmark datasets. For instance, the model's proficiency in medical tasks demonstrates superior performance on datasets such as MedQA and PubMedQA compared to previous models like PubMedGPT and BioGPT. Moreover, Li et al.~\cite{Li:2023} developed ChatDoctor, based on LLaMA, emphasizing the importance of fine-tuning LLMs with domain-specific data to enhance their performance and reliability in medical contexts. By training on a substantial dataset composed of $100,000$ doctor-patient interactions and incorporating real-time information retrieval mechanisms from online sources like Wikipedia, ChatDoctor exemplifies a significant leap toward creating autonomous, knowledgeable medical chatbots.

Furthermore, a recent study proposed by Mehandru et al.~\cite{Mehandru:2024} explores the integration of LLMs in clinical settings. The research examines the practical application of LLMs as clinical agents, focusing on their ability to support healthcare professionals by providing reliable and relevant medical information.

This paper proposes fine-tuning the ChatBode-7B model using datasets translated into Brazilian Portuguese to develop a virtual medical assistant, specifically a chatbot specializing in medicine. \red{Ideally, the model would be fine-tuned using a native dataset consisting of verified medical conversations; however, such a dataset is currently unavailable in the literature. Existing datasets either consist of historical medical conversations from the 16th century~\cite{zilio2024nlp} or are translations that lack professional verification~\cite{askdbr2022gomes}.} To address this, we created a Portuguese corpus using datasets such as HealthCareMagic-100k-en and MedQuAD, which were translated using GPT-3.5. The fine-tuning process employs the PEFT-QLoRA method, combining various medical and instruction-following data to enhance the model's ability to generate accurate and relevant medical responses.

The study aims to create a robust medical chatbot in Portuguese to improve access to medical information for Portuguese-speaking populations, enhance healthcare outcomes, and address these communities' specific linguistic and cultural needs. Additionally, it is worth noting that, until the date of writing this article, no medical assistant in Portuguese had been found in the literature, making this work pioneering.

The remainder of this paper is presented as follows. \red{Section~\ref{s.methodology} describes the proposed methodology regarding the development stages of DrBode, including data preparation and the systematic experimentation involved. Section~\ref{s.results} presents and discusses the experimental results, focusing on regional factors and the associated risks of applying generative AI in medicine. Lastly, Section~\ref{s.conclusions} provides the study's conclusions and suggests future research directions.}


\section{Methodology}
\label{s.methodology}

This section details the development stages of DrBode, illustrating the rigorous processes and methods employed in its creation. The methodology encompasses meticulous data preparation and systematic experimentation to assess model performance across various configurations.

\subsection{Dataset}

\red{
Due to the lack of native Brazilian Portuguese datasets focused on medical conversations, we translated healthcare-related datasets into Brazilian Portuguese as a temporary solution. While this approach facilitates fine-tuning the model for the medical domain, it does not fully address critical issues such as cultural diseases and specific healthcare aspects relevant to the Brazilian population. Nevertheless, we ensured terminological and semantic accuracy during the translation to mitigate potential discrepancies. The datasets used are:
}

\begin{itemize}
    \item HealthCareMagic-100k-en\footnote{https://huggingface.co/datasets/wangrongsheng/HealthCareMagic-100k-en}: This dataset consists of approximately $100,000$ samples of doctor-patient interactions, originally in English, used in the ChatDoctor~\cite{}. The translation into Portuguese was carried out using the GPT-3.5 model, ensuring terminological and semantic consistency.
    \item MedQuAD\footnote{https://huggingface.co/datasets/lavita/MedQuAD}: This dataset comprises approximately $50,000$ sample question-and-answer pairs related to the medical field. However, only about $9,500$ samples were utilized because complete answers were unavailable. This dataset was crucial in broadening the range of medical scenarios examined during fine-tuning.
\end{itemize}
    
\red{Although these datasets aligned well with the medical domain and Portuguese language. A comprehensive, culturally relevant dataset is still needed to address the nuances of Brazilian healthcare, including diseases specific to the region and the linguistic particularities of Brazilian Portuguese. We hope that this work will push the field forward and showcase the necessity of developing native datasets that better reflect the linguistic and cultural nuances of Brazilian healthcare, enabling more accurate and context-aware medical AI models.
}

\subsection{Fine-Tuning}

The fine-tuning process of the InternLM2-chatbode-7b 
model was conducted following the QLoRA methodology~\cite{dettmers2023qlora}, focused on adapting the model to effectively respond to medical queries in Brazilian Portuguese. This method incorporates a series of strategic steps to optimize the model's specialization and generalization capabilities. The QLoRA methodology encompasses three distinct approaches for model fine-tuning, being:
\begin{itemize}
    \item Fine-tuning exclusively with medical data: The model was trained exclusively with data from the HealthCareMagic-100k-en and MedQuAD datasets, aiming to maximize medical specialization;
    \item Fine-tuning with 50\% medical data and 50\% instruction-following data: Half of the medical data was combined with samples from different tasks, included from the openHermes\footnote{https://huggingface.co/datasets/adalbertojunior/openHermes\_portuguese} and Dolphin Portuguese\footnote{https://huggingface.co/datasets/adalbertojunior/dolphin\_portuguese} datasets.
     This approach aims to balance medical expertise with the ability to follow instructions in varied contexts;
    \item Fine-tuning with 33\% medical data and 67\% instruction-following: A larger proportion of task-diverse data was introduced to explore the adaptability of the model in non-medical contexts, while still benefiting from the specific knowledge gained from medical data.
\end{itemize}

Notice that the instruction-following samples sourced from openHermes\_portuguese and dolphin\_portuguese datasets were selected due to the high performance of models trained with these data on the Open PT LLM leaderboard.

\subsection{Question Evaluation}
\red{As there is no current objective benchmark available that can effectively evaluate cultural nuances specific to Brazilian Portuguese, nor is there a well-established framework for the objective evaluation of models in this particular scenario, we adopted a qualitative evaluation method supported by medical professionals. For this study, eight medical professionals evaluated five critical queries, each judged based on six distinct criteria to analyze the geocultural aspects of medical responses. This process resulted in a total of 240 evaluations, providing a comprehensive analysis that reflects real-world medical concerns and the cultural specificity of patient care in Brazil. The questions can be observed as follows:}

\begin{itemize}
    \item The doctor prescribed atenolol for high blood pressure, but I don't have high blood pressure, just palpitations from anxiety. Can atenolol be used for anxiety?
    \item My period is late, and I read that cinnamon tea can make it come faster. Is it possible to induce menstruation in 30 minutes?
    \item I have mouth ulcers, and a friend suggested nystatin oral. What is it for, and how should it be used? Is it indicated for mouth ulcers?
    \item I have body aches, a headache, high fever, and extreme fatigue. What could it be? Should I take any medication?
    \item Explain in detail the difference between the flu and a cold.
\end{itemize}

We assessed responses from various models: the base model InternLM2\footnote{https://huggingface.co/internlm/internlm2-chat-7b}, the ChatBode 
\red{which} consists of InternLM2 fine-tuned with Ultra Alpaca without healthcare data and the two models fine-tuned with different proportions of medical and instruction-following data.

The qualitative assessment of the model responses was conducted by medical professionals, considering the critical nature of accuracy in medical advice. The evaluation was necessary because quantitative methods for text generation in Portuguese are still scarce, particularly in the medical domain where issues like hallucinations in generated text can have significant negative impacts. Evaluating by specialists helped identify the potential beneficial uses of these models while highlighting necessary cautions and alerts.

Responses generated by the model trained solely with medical data were excluded from the evaluation due to significant issues in response formulation, being evidently worse than those generated by other models. This exclusion and its implications are further discussed in Section~\ref{s.results}.

The evaluation criteria were: 
\begin{itemize}
    \item Accuracy (0-5): Correctness of information.
    \item Completeness (0-5): Thoroughness of the response.
    \item Adequacy (0-5): Appropriateness of tone and style.
    \item Safety (0-5): Potential health risks posed by the response.
    \item Grammaticality (0-5): Adherence to the standard Portuguese language.
    \item Coherence (0-5): Logical flow and structure of the response.
\end{itemize}

This approach ensured a thorough evaluation of the models' potential benefits and risks in providing medical information. Additionally, evaluators had the option to provide further comments to elaborate on their ratings, ensuring a detailed and nuanced evaluation.
	
\section{Results and Discussion}
\label{s.results}

This study sought to address several critical aspects of the application of generative Large Language Models (LLMs) in the medical domain. Among the models evaluated, as the results in Table~\ref{tab:evaluation_metrics} show, InternLM2 demonstrated superior performance across several metrics. This superiority likely stems from their initial training phase, which appears to have incorporated medical data, as indicated by other studies~\cite{qiu2024mmedlm}. 

\begin{table}[h!]
\centering
\caption{Evaluation Metrics for Different Models}
\label{tab:evaluation_metrics}
\resizebox{\linewidth}{!}{
\begin{tabular}{lcccc}
\toprule
\textbf{Criterion} & \textbf{InternLM2} & \textbf{ChatBode} & \textbf{DrBode 360} & \textbf{DrBode 240} \\
\midrule
\textbf{Accuracy} & 3.8 $\pm$ 1.4 & 3.1 $\pm$ 1.2 & 3.6 $\pm$ 1.3 & 3.4 $\pm$ 1.2 \\
\textbf{Completeness} & 3.7 $\pm$ 1.5 & 3.0 $\pm$ 1.0 & 3.3 $\pm$ 1.3 & 3.4 $\pm$ 1.0 \\
\textbf{Adequacy} & 3.7 $\pm$ 1.6 & 3.2 $\pm$ 1.1 & 3.3 $\pm$ 1.4 & 3.5 $\pm$ 1.0 \\
\textbf{Safety} & 4.0 $\pm$ 1.3 & 3.5 $\pm$ 1.2 & 3.3 $\pm$ 1.6 & 3.2 $\pm$ 1.3 \\
\textbf{Grammaticality} & 3.9 $\pm$ 1.5 & 4.3 $\pm$ 0.8 & 4.2 $\pm$ 0.9 & 3.8 $\pm$ 1.1 \\
\textbf{Coherence} & 4.1 $\pm$ 1.4 & 4.3 $\pm$ 0.8 & 4.2 $\pm$ 0.8 & 4.2 $\pm$ 0.7 \\
\bottomrule
\end{tabular}
}
\end{table}


\red{In contrast, the DrBode models, which were fine-tuned from ChatBode—a version of InternLM2 adapted for instruction-following in Portuguese but lacking medical-specific data—suffered from catastrophic forgetting. This phenomenon occurs when a model loses previously learned knowledge while acquiring new information, which, in this case, results in a decline in medical domain performance. Although DrBode demonstrated improvements over ChatBode, it could not match the performance of InternLM2, particularly regarding medical accuracy and reliability. The fine-tuning process on medical data appeared to overwrite key general knowledge learned by InternLM2, causing a degradation in its ability to deliver high-quality medical responses.}

\begin{figure}[h]
\includegraphics[width=0.475\textwidth]{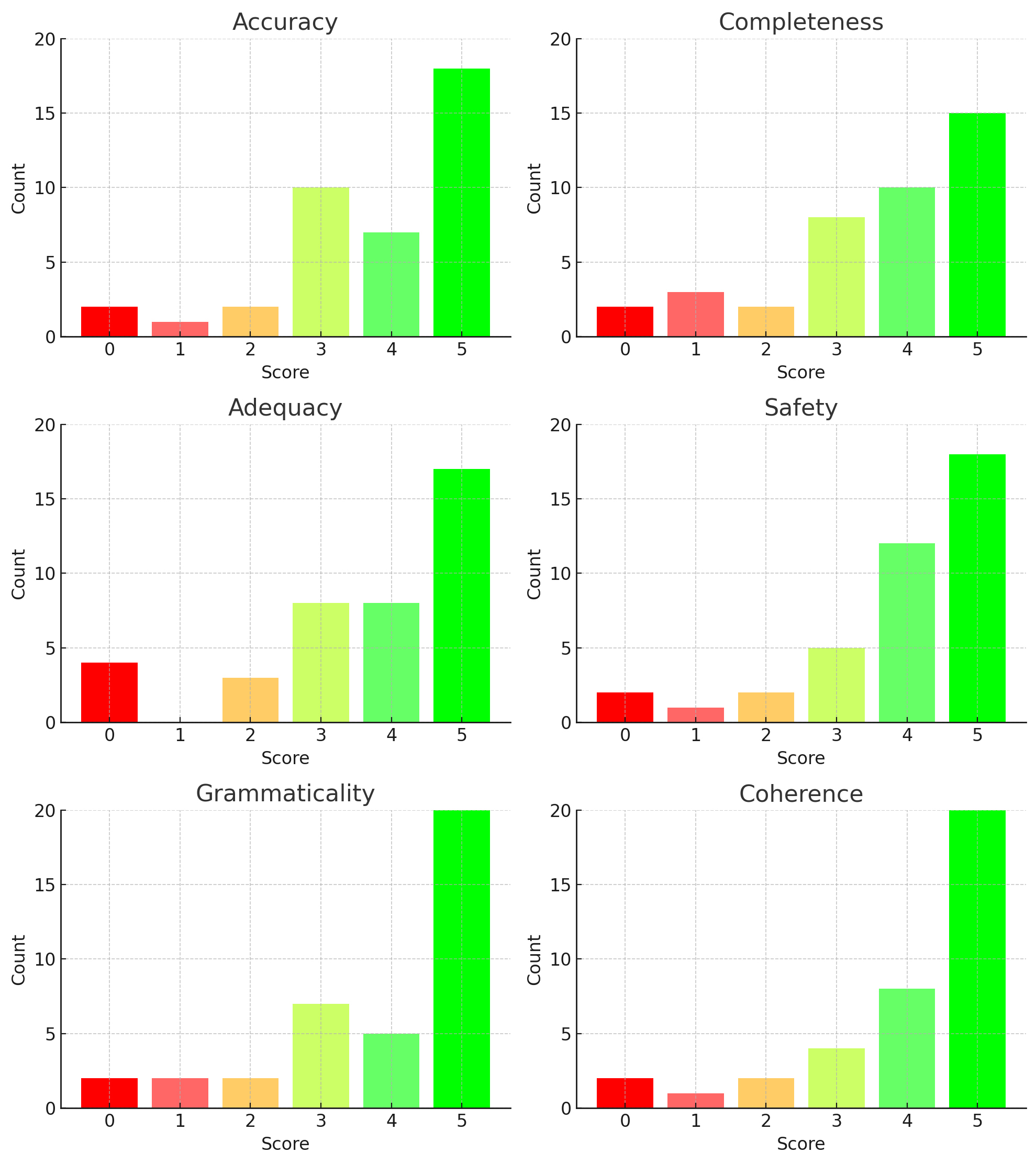}
\caption{Distribution of evaluator scores for the InternLM2 model.}
\label{fig:InternLM2}
\end{figure}

\begin{figure}[h]
\includegraphics[width=0.475\textwidth]{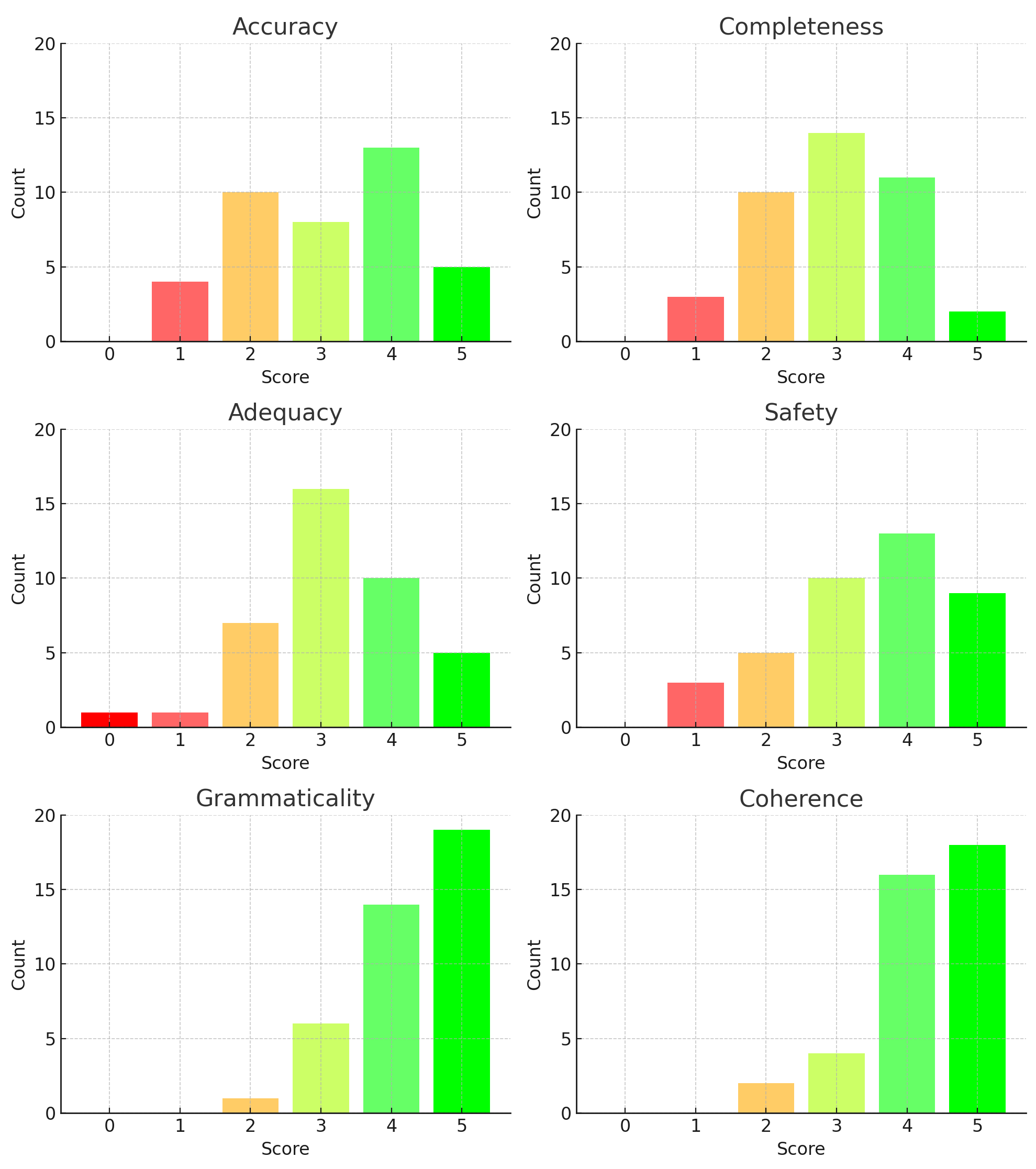}
\caption{Distribution of evaluator scores for the ChatBode model.}
\label{fig:chatbode}
\end{figure}

\begin{figure}[h]
\includegraphics[width=0.475\textwidth]{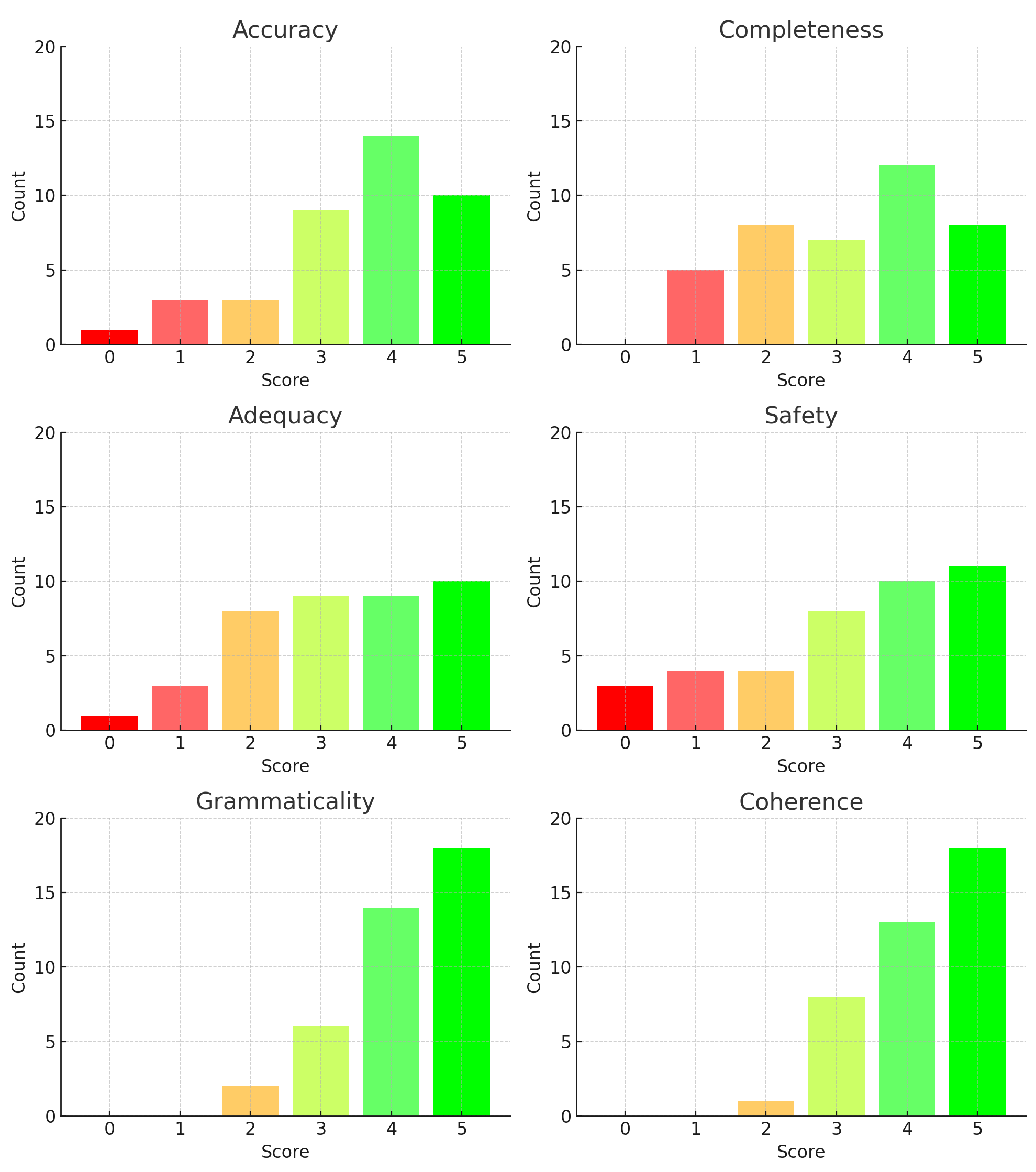}
\caption{Distribution of evaluator scores for the DrBode 360 model.}
\label{fig:DrBode360}
\end{figure}

\begin{figure}[h]
\includegraphics[width=0.475\textwidth]{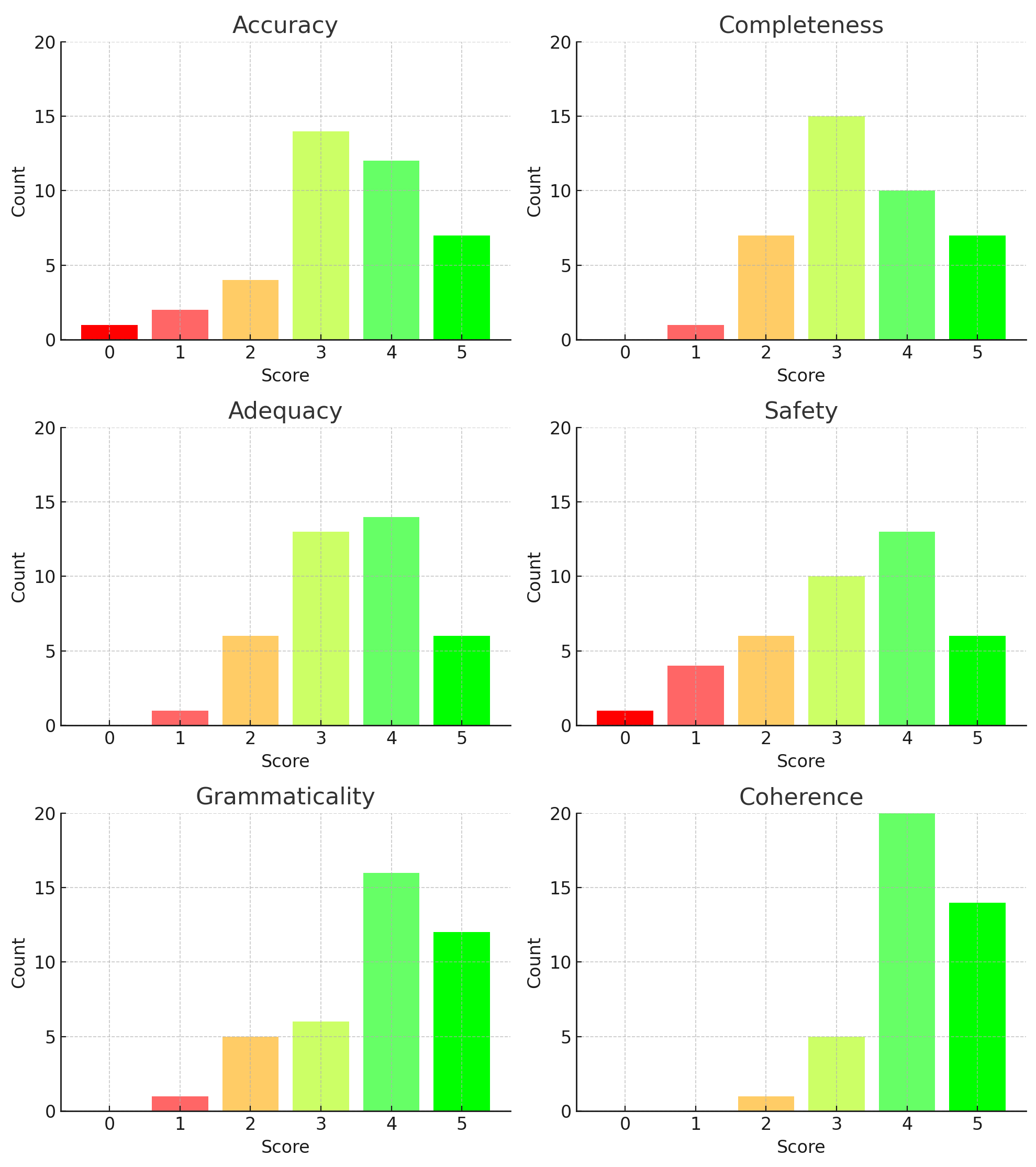}
\caption{Distribution of evaluator scores for the DrBode 240 model.}
\label{fig:DrBode240}
\end{figure}

Despite InternLM2's overall dominance, it is noteworthy that in terms of grammaticality and coherence, the other models performed comparably or even slightly better. Furthermore, these models generally exhibited lower standard deviations in their performance metrics, suggesting a higher level of consistency in their responses. This aspect of performance underscores the potential of these models to deliver reliable and grammatically coherent responses, which are crucial for maintaining the professionalism and clarity required in medical communication.

The distribution of evaluators' responses for the InternLM2, ChatBode, DrBode 360, and DrBode 240 models, presented in Figures~\ref{fig:InternLM2},~\ref{fig:chatbode},~\ref{fig:DrBode360}, and~\ref{fig:DrBode240} respectively, provides further insight into the models' performance. Notably, these figures illustrate that InternLM2, despite its overall superior performance, received a higher number of 0 scores across most criteria compared to the other models. This indicates areas where even the most robust model may fail, highlighting the importance of continual refinement and validation in diverse real-world scenarios.

An important factor to consider when evaluating these models based on the scores assigned by the evaluators is the degree of agreement among them. To assess this, Kendall's coefficient was calculated for each evaluated criterion, both by model (Table~\ref{tab:kendall_coefficient}) and by question (Table~\ref{tab:kendall_coefficient_questions}). Overall, the results indicate a low degree of agreement in most scenarios, highlighting the challenges in achieving consistent evaluations in this context.

\begin{table}[h!]
\centering
\caption{Kendall's Coefficient of Concordance for Different Models}
\label{tab:kendall_coefficient}
\resizebox{\linewidth}{!}{
\begin{tabular}{lcccc}
\toprule
\textbf{Criterion} & \textbf{InternLM2} & \textbf{ChatBode} & \textbf{DrBode 360} & \textbf{DrBode 240} \\
\midrule
{Accuracy} & \cellcolor{green!25}0.672 & \cellcolor{red!25}0.109 & \cellcolor{red!25}0.004 & \cellcolor{yellow!25}0.305 \\
{Completeness} & \cellcolor{green!25}0.689 & \cellcolor{red!25}0.035 & \cellcolor{red!25}0.241 & \cellcolor{yellow!25}0.377 \\
{Adequacy} & \cellcolor{green!25}0.501 & \cellcolor{red!25}0.150 & \cellcolor{red!25}0.167 & \cellcolor{red!25}0.225 \\
{Safety} & \cellcolor{green!25}0.678 & \cellcolor{red!25}0.103 & \cellcolor{red!25}0.202 & \cellcolor{red!25}0.284 \\
{Grammaticality} & \cellcolor{green!25}0.685 & \cellcolor{red!25}0.276 & \cellcolor{red!25}0.186 & \cellcolor{red!25}0.262 \\
{Coherence} & \cellcolor{green!50}0.711 & \cellcolor{red!25}0.256 & \cellcolor{red!25}0.214 & \cellcolor{red!25}0.107 \\
\bottomrule
\end{tabular}
}
\end{table}


\begin{table}[h!]
\centering
\caption{Kendall's Coefficient of Concordance for Each Question}
\label{tab:kendall_coefficient_questions}
\begin{tabular}{lccccc}
\toprule
\textbf{Question} & \textbf{InternLM2} & \textbf{ChatBode} & \textbf{DrBode 360} & \textbf{DrBode 240} \\
\midrule
{1} & \cellcolor{green!25}0.558 & \cellcolor{yellow!25}0.497 & \cellcolor{yellow!25}0.460 & \cellcolor{red!25}0.174 \\
{2} & \cellcolor{red!25}0.112 & \cellcolor{yellow!25}0.331 & \cellcolor{green!25}0.525 & \cellcolor{red!25}0.258 \\
{3} & \cellcolor{red!25}0.159 & \cellcolor{yellow!25}0.395 & \cellcolor{yellow!25}0.326 & \cellcolor{red!25}0.207 \\
{4} & \cellcolor{red!25}0.110 & \cellcolor{yellow!25}0.393 & \cellcolor{yellow!25}0.318 & \cellcolor{yellow!25}0.468 \\
{5} & \cellcolor{red!25}0.012 & \cellcolor{yellow!25}0.428 & \cellcolor{red!25}0.161 & \cellcolor{red!25}0.291 \\
\bottomrule
\end{tabular}
\end{table}

This low level of agreement between evaluators is a major complication for evaluating the quality of model responses. The evaluators consistently displayed a high degree of variability in their ratings, pointing to an inherent complexity in evaluating AI-generated content in specialized domains such as healthcare. The reasons for this variability are not entirely clear and represent a critical area for further investigation. Future studies might benefit from a more structured evaluation framework or more rigorous training for evaluators to ensure a higher consistency and reliability in ratings.

The observed discrepancies and the challenge of evaluator concordance highlight the need for ongoing research into the application of LLMs in healthcare. This research should prioritize not only the technical refinement of models to prevent knowledge loss but also the development of evaluation methodologies that can more accurately reflect the utility and safety of AI applications in sensitive and high-stakes fields like medicine.

\subsection{Regional Factors and the Risks of Generative AI in Medicine}

The deployment of generative artificial intelligence (AI) in healthcare introduces a complex array of ethical, moral, and legal implications that remain largely unresolved. The potential misuse of these models can exacerbate existing health risks associated with self-diagnosis and self-medication—practices that are already prevalent due to the ease of searching for symptoms, diseases, and treatments online. Large Language Models (LLMs) further complicate this landscape due to their inherent capacity to generate hallucinated or inaccurate information.

This research acknowledges these challenges, and as a precaution, the trained models will be distributed with a usage license that highlights these risks. During our study, healthcare professionals who evaluated the model responses noted a concerning tendency: the models frequently suggested pharmaceutical interventions. \red{A striking example is the case of dengue, for instance, in response to the query:} "I have body aches, a headache, high fever, and severe fatigue. What could it be? Should I take any medication?", the model DrBode 240 suggested taking ibuprofen to alleviate symptoms. \red{However, such symptoms are indicative of dengue fever, a disease endemic to Brazil. In cases of dengue, the use of non-steroidal anti-inflammatory drugs (NSAIDs) like ibuprofen can lead to severe complications such as internal bleeding, dramatically worsening the patient's condition and significantly increasing the risk of mortality.}

Furthermore, this issue highlights a significant concern regarding the data on which these models are trained. Predominantly, the models are not trained on data originally in Portuguese but rather on translated datasets. This approach can lead to a model's failure to capture and consider regional nuances in its responses. For example, while all models could identify dengue when asked directly about the disease, none considered it as a potential diagnosis in response to the symptoms described above. For regions where dengue is endemic, such as Brazil, it would be natural to consider it as a primary possibility based on the presented symptoms. \red{This is a critical scenario that emphasizes the urgent necessity of acquiring native datasets that accurately reflect regional health challenges, ensuring models are equipped to handle culturally specific medical issues effectively.}

In conclusion, while generative AI can significantly enhance informational accessibility and decision-making in healthcare, it is crucial to approach its integration with caution, ensuring that models are not only accurate but also sensitive to the regional and cultural contexts in which they are deployed. This requires continuous evaluation and adaptation of the models to mitigate risks and enhance their reliability and applicability in diverse healthcare environments.

\section{Conclusions and Future Works}
\label{s.conclusions}

This study has evaluated the performance of generative Large Language Models (LLMs) within the medical domain, revealing distinct capabilities and limitations. Among the models assessed, InternLM2 emerged as the top performer, likely benefiting from its initial training that included medical datasets. This prior exposure to medical content likely equipped InternLM2 with a nuanced understanding and accuracy in medical contexts, as evidenced by its superior performance in most evaluated metrics.

On the other hand, the DrBode models—derived from iterative fine-tuning of ChatBode, which itself builds on InternLM2—\red{suffered from catastrophic forgetting from the original InternLM model, particularly in medical knowledge. The absence of medical-specific data during ChatBode's instruction-following fine-tuning in Portuguese contributed to this degradation. This helps us emphasize the challenge of maintaining domain-specific expertise through successive fine-tuning stages. Although DrBode models exhibited improvements over ChatBode, they failed to reach the benchmark set by InternLM2 in terms of medical accuracy.}

It is important to note, however, that in areas such as grammaticality and coherence, the DrBode models performed comparably to or even surpassed InternLM2. These models also demonstrated greater consistency in their outputs as reflected by generally lower standard deviations across most metrics. Such attributes highlight their potential reliability and usefulness in scenarios where linguistic precision and consistency are paramount.

\red{A critical finding of this study is the urgent need to improve fine-tuning processes. In future works, we aim to evaluate the direct Fine-Tuning of the InternLM2 model and improve the data composition of the dataset in order to mitigate the catastrophic forgetting observed in DrBode in relation to the original InternLM2 model by ensuring that medical knowledge is not overwritten during fine-tuning. Directly fine-tuning InternLM2 on medical data rather than using intermediary models like ChatBode may preserve the strong foundational knowledge critical for medical applications.}

\red{
Furthermore, this research revealed the urgent necessity for native datasets in less-resourced languages such as Brazilian Portuguese. Current datasets rely heavily on translations, which do not account for critical cultural nuances, leading to potential risks in clinical settings. For instance, in handling regional cases like Dengue, LLMs prescribed dangerous medications due to a lack of localized medical knowledge. The development of native datasets that accurately represent regional health issues is essential for addressing these gaps.
\\
This study also uncovered a significant gap in the literature, where even powerful models like InternLM2 struggle to capture cultural nuances properly. Despite their capabilities, these models still fail to represent specific regional medical contexts and culturally relevant health conditions, a crucial area that remains underexplored.
}

Another significant challenge identified through this research was the low concordance among evaluators, which underscores the complexities involved in assessing AI-generated content in specialized domains. This variability in evaluations suggests that standardizing assessment protocols or enhancing evaluator training could be crucial in future research endeavors.

The findings of this study pave the way for several future research directions:

\begin{itemize}
    \item \textbf{Evaluating Multilingual Models:} The recent introduction of models like MMedLM~\cite{qiu2024mmedlm}, designed specifically for multilingual medical applications, presents an opportunity to evaluate and possibly enhance the performance of medical LLMs in languages other than English, particularly in Portuguese. Future work should focus on assessing these models' effectiveness and conducting targeted fine-tuning to cater to regional medical needs.

    \item \textbf{Developing Native Datasets:} There is a critical need to refine the datasets used for training these models by incorporating native language data that reflect specific regional health scenarios, such as endemic diseases in Brazil. This approach would likely improve the models' contextual relevance and accuracy in local settings.

    \item \textbf{Improving Evaluation Consistency:} Addressing the low evaluator concordance observed in this study is imperative, \red{with the lack of a current robust benchmark for clinical validity of the models that doesn't rely only on qualitative evaluation}. Future research should explore more robust evaluation frameworks or methodologies that can reduce subjectivity and enhance the reliability of model assessments.

    \item Integrating External Knowledge Sources: To mitigate issues related to inaccurate or hallucinated content, integrating external, verified medical knowledge bases could enhance the reliability and safety of AI-generated advice in healthcare settings.
\end{itemize}

By addressing these areas, future research can not only refine the utility of LLMs in healthcare but also ensure their ethical and safe integration into medical practice. This would align technological advancements with the overarching goal of improving patient care and public health outcomes.



\bibliographystyle{IEEEtran}
\bibliography{ref}

\end{document}